\documentclass[journal,twoside,web]{ieeecolor}
\usepackage{generic}
\usepackage{cite}
\usepackage{amsmath,amssymb,amsfonts}
\usepackage{graphicx}
\usepackage{textcomp}
\usepackage[hidelinks]{hyperref}
\hypersetup{
  pdftitle={Towards Clinically Faithful Medical Image Captioning via Enhanced Vision-Language Alignment},
  pdfauthor={Yunseo Lee, Hyun Jun Kim, Heeseung Shin, and Changwon Lim},
  pdfsubject={Preprint submitted to IEEE for possible publication},
  pdfkeywords={medical image captioning, clinical alignment, multimodal learning, reinforcement learning, vision-language model}
}
\def\BibTeX{{\rm B\kern-.05em{\sc i\kern-.025em b}\kern-.08em
    T\kern-.1667em\lower.7ex\hbox{E}\kern-.125emX}}
\makeatletter
\def\ps@titlepagestyle{%
  \def\@oddhead{\scriptsize\textsf{PREPRINT}\hfill\thepage}%
  \def\@evenhead{\scriptsize\thepage\hfill\textsf{PREPRINT}}%
  \def\@oddfoot{}\def\@evenfoot{}}
\makeatother

\begin{document}
\title{Towards Clinically Faithful Medical Image Captioning via Enhanced Vision–Language Alignment}

\author{Yunseo Lee, Hyun Jun Kim, Heeseung Shin, and Changwon Lim%
\thanks{Yunseo Lee is with the Department of Statistics and Data Science, Chung-Ang University, 84 Heukseok-ro, Dongjak-gu, Seoul 06974, Republic of Korea (e-mail: 34046a@gmail.com).}
\thanks{Hyun Jun Kim is with the Department of Smart Cities, Chung-Ang University, 84 Heukseok-ro, Dongjak-gu, Seoul 06974, Republic of Korea (e-mail: hyunjun0615@cau.ac.kr).}
\thanks{Heeseung Shin is with the Department of Statistics and Data Science, Chung-Ang University, 84 Heukseok-ro, Dongjak-gu, Seoul 06974, Republic of Korea (e-mail: hs970416@cau.ac.kr).}
\thanks{Changwon Lim is with the Department of Applied Statistics, Chung-Ang University, 84 Heukseok-ro, Dongjak-gu, Seoul 06974, Republic of Korea, and is the corresponding author (e-mail: clim@cau.ac.kr).}
}

\maketitle

\begin{center}
\small\emph{This work has been submitted to the IEEE for possible publication. Copyright may be transferred without notice, after which this version may no longer be accessible.}
\end{center}

\begin{abstract}
Medical image captioning is an important technique that can accelerate early-stage diagnostic workflows and enhance the interpretability of medical diagnostic AI systems. However, unlike general image captioning, generating clinically reliable captions remains challenging due to grayscale-based modalities, subtle anatomical cues, specialized medical phrasing, and variations in data quality. Despite recent advances in large vision--language models, fluent outputs do not necessarily guarantee sufficient alignment with clinical concept spaces or evaluation criteria.
To address this issue, we propose a framework that strengthens clinical alignment by separating and enhancing training-time alignment and inference-time alignment. We build a medical image captioning pipeline that integrates single/dual vision encoders based on BioMedCLIP and SigLIP2, a Q-Former, and a LLaMA-based decoder, and examine the contribution of auxiliary learning for UMLS concept/type prediction. At inference, we apply single-embedding-based reranking to select the most appropriate caption among candidates, while at training we introduce MedPAIR-SCST, which combines clinically relevant rewards to shift the generative distribution toward improved clinical alignment.
Our experiments show that leveraging complementary visual representations with a multi-encoder design and concept-level auxiliary learning can contribute to preserving clinically meaningful information. Furthermore, inference-time reranking provides a practical way to improve semantic and clinical alignment without additional training, whereas MedPAIR-SCST goes beyond selection by directly improving the model's distribution to generate more consistent and clinically grounded captions. These findings suggest that jointly leveraging selection-based alignment and reinforcement-learning-based alignment can promote more trustworthy medical image captioning even in data-constrained settings.
\end{abstract}


\section{Introduction}
\label{sec:introduction}
\IEEEPARstart{M}{edical} image captioning---automatically generating radiologist-style descriptions from imaging studies---has the potential to accelerate report drafting, improve content-based image retrieval, and increase the interpretability of diagnostic AI models. Compared with natural-image captioning, the task is complicated by grayscale modalities, subtle anatomical cues, and a highly specialized vocabulary, all of which demand fine-grained visual reasoning and domain knowledge \cite{1}. While recent systems have made notable progress with encoder--decoder frameworks trained on paired image--text datasets, performance remains constrained by data quality, domain adaptability, and output reliability \cite{2},\cite{3},\cite{4}. Public medical corpora frequently contain low-resolution images \cite{5} and annotation-induced artifacts \cite{6}, degrading perception and propagating spurious patterns into generation. Moreover, generic vision encoders may miss subtle, domain-specific signals \cite{7}, and caption decoders can produce inconsistent or incomplete statements due to limited grounding in clinical semantics \cite{8}.

Despite advances in large vision--language models (VLMs), two structural gaps persist in practice. First, general-purpose pretraining induces a global-semantics bias that under-attends cues dispersed over small receptive fields, encouraging formulaic phrasing (e.g., ``mild cardiomegaly,'' ``no acute process''). Second, there is no guarantee that generated text is automatically aligned with clinical concept spaces (terminologies/types) or with evaluation metrics (e.g., BERTScore, ROUGE, biomedical similarity). Outputs may read naturally yet remain weakly supported or terminologically inconsistent, motivating a shift from fluency-centric modeling to clinical alignment as a primary objective.

We address this gap by disentangling learning from inference and evaluating them separately. As a Stage-1, we train MedCap under six configurations to establish a reproducible baseline for comparison. This allows us to analyze how performance changes with encoder composition and auxiliary supervision, and to identify potentially complementary behaviors across models.

At inference time, we incorporate single-embedding based reranking as an alignment method (not a learning stage): among candidate captions produced by the Stage-1 models, we select the sentence with the highest image--text compatibility under a pre-specified embedding model. This post-hoc selection requires no parameter updates or additional data, directly injects a vision--language alignment signal that tends to reduce hallucinations and improve relevance, and admits simple policy controls (e.g., thresholding, length normalization) for deployment.

However, post-hoc selection does not modify the model's generative distribution. We therefore introduce Stage-2 (training-time alignment) via a modified self-critical sequence training procedure. For each input, multiple candidates are sampled and re-scored under teacher-forcing sequence log-likelihoods; optimization then shifts the distribution toward clinically aligned metrics. Our modification emphasizes practical stability without a reference model or KL regularization by combining group-level reward aggregation with reference-free pairwise comparison, mitigating the high-variance behavior commonly observed in standard SCST. In short, inference-time reranking selects better candidates, whereas Stage-2 improves the generator itself; the two are complementary.

Under identical decoding, we compare Stage-1 SFT models, inference-time alignment methods at test time (reranking), and Stage-2 training based on a modified self-critical objective. We report textual metrics (BERTScore, ROUGE, BLEURT) and biomedical similarity measures, and we monitor reward--metric alignment during training to verify that improvements follow the intended evaluation direction rather than superficial overfitting. Implementation details (ensemble and selection policies) are deferred to the Appendix; the main text focuses on the principles and effects of alignment.

The contributions of this work are as follows:
\renewcommand{\labelenumi}{\arabic{enumi})}
\begin{enumerate}
    \item Constructed an end-to-end framework for medical image captioning by integrating multi-encoder visual grounding, a query-aggregation module, and a Llama-based decoder with supervised fine-tuning.
    \item Strengthened clinical alignment along two orthogonal axes rather than a sequential pipeline: (a) inference-time selection via post-hoc reranking and (b) training-time distributional optimization via an enhanced self-critical objective; each axis is evaluated independently and can be applied on its own.
    \item Proposed a reference/KL-free self-critical variant that combines group-level reward summarization with reference-free pairwise ordering, yielding clinical alignment gains without a reference model or KL regularization.
\end{enumerate}

\section{Related work}
 Medical image captioning has evolved alongside advances in vision-language modeling, primarily following the encoder-decoder paradigm widely used in natural image captioning. Early works employed convolutional neural networks (CNNs) as visual encoders paired with recurrent neural networks (RNNs) or Transformer-based decoders to generate captions \cite{1}. However, these approaches often lacked clinical specificity, as they relied on general-purpose image features and were trained on limited or noisy medical datasets.
 
Medical image encoders are now designed primarily within a vision–language pretraining paradigm that goes beyond single-modality feature extraction to jointly model images and their accompanying reports. BioViL-T \cite{50} extends this approach with a hybrid CNN–Transformer encoder for longitudinal chest X-rays and associated reports. BiomedCLIP \cite{12} builds a CLIP-style biomedical foundation model by pretraining a ViT vision encoder and PubMedBERT text encoder on PMC-15M, a dataset of over 15 million figure–caption pairs. By learning domain-specific image–text embeddings from large-scale medical corpora, these encoders provide clinically aware visual representations that substantially improve transfer to downstream classification, retrieval, and reporting tasks over generic natural-image encoders.

On the decoder side, medical-domain large language models (LLMs) have been proposed as specialized generators for biomedical and clinical text. BioGPT \cite{51} is a generative Transformer pretrained on large-scale PubMed abstracts, achieving state-of-the-art performance on multiple biomedical NLP benchmarks. Med-PaLM \cite{52} adapts a general-purpose LLM (Flan-PaLM) to the medical domain via instruction tuning on MultiMedQA, reaching near-clinician-level accuracy on exam-style questions. These medical-domain decoders provide a natural generator for captioning and report-generation systems that require both accurate biomedical terminology and clinically appropriate reasoning.

Beyond using medical encoders and decoders in isolation, several recent systems propose full medical vision–language models (VLMs) that jointly process images and text and can generate free-form clinical descriptions. LLaVA-Med \cite{53} extends a general vision–language assistant (LLaVA) to the biomedical domain via two-stage multimodal instruction tuning on PMC-15M figure–caption pairs and GPT-4–generated biomedical conversations. XrayGPT \cite{54} specializes this paradigm to chest radiographs by combining a MedCLIP visual encoder with a Vicuna-based medical LLM through a learnable projection layer to support radiology-report summarization and interactive question answering over X-rays. These medical VLMs show that pairing domain-specific visual encoders with instruction-tuned medical (or general) LLMs can substantially improve clinical faithfulness and versatility in medical image captioning and reporting.

Across image, video, and audio captioning, a common post-hoc strategy is to first produce multiple candidate captions and then rerank them using a compatibility score computed in a shared multimodal embedding space. Fang et al. (2015) \cite{38} captioning system generates many candidate sentences with a maximum entropy language model and then applies a Deep Multimodal Similarity Model together with sentence-level features to rerank the hypotheses, thereby operationalizing selection in an image–text semantic space. In the video domain, BLIP4video \cite{40} first generates multiple candidate captions and then reranks them using both a video–text alignment score and a CIDEr-based n-gram similarity. Subsequently, two lightweight post-hoc procedures apply these scores at intra- and inter-inference stages, and the final caption is selected from the reordered set. SLAM-AAC \cite{42} introduces CLAP-Refine as a plug-and-play inference step that computes audio–text similarity in a CLAP embedding space and reranks beam-search outputs to pick the final caption. Complementary to reranking, an alternative post-hoc selection paradigm uses a large language model to summarize the set of candidate captions into a single, comprehensive description—for example, IC3 \cite{43} samples diverse captions from a base captioner and prompts a vision-free LLM to synthesize them into one consolidated caption.

If reranking is a test-time post-hoc procedure that selects the best sentence from a pool of hypotheses, Self-Critical Sequence Training (SCST) \cite{44} internalizes the same selection criteria into the training objective so that the generator learns to produce better sentences outright. Building on this classical line of work, recent work in reinforcement and preference learning broadens the selection signal. PPO \cite{45} has been applied to image-to-text generation as a trust-region policy-gradient alternative that controls policy shift and KL divergence while stabilizing CIDEr-driven optimization. DPO \cite{46} subsequently emerged as a reference-light preference optimizer; in the clinical setting, RRG-DPO \cite{47} curates preferred/dispreferred/sub-preferred pairs using biomedical-CLIP neighborhoods and abnormality conflicts, aligning report generation on X-ray and CT toward clinically faithful preferences. Finally, GRPO \cite{48} introduces group-wise relative optimization to captioning: the model samples multiple hypotheses per image and maximizes within-group relative advantages under a KL constraint, reducing update variance and jointly improving stability and diversity relative to SCST.

\section{Method}
\subsection{Base Model Architecture}
\begin{figure}[ht]
    \centering
    \includegraphics[width=0.9\linewidth]{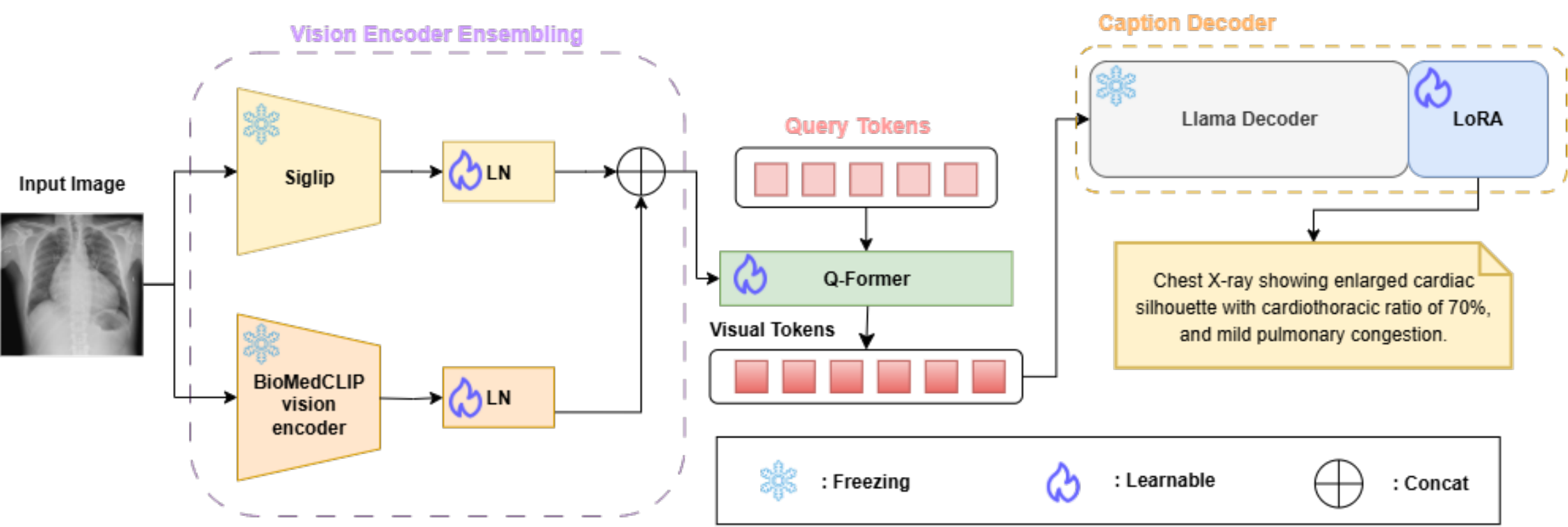}
    \caption{The figure illustrates the architecture of a medical image captioning model that generates a final caption by fusing outputs from two vision encoders, followed by a Q-Former and a LLaMA decoder.; CC BY-NC [Al Mulhim et al.] \cite{36}}
    \label{fig:architecture}
\end{figure}

The overall architecture of our proposed medical image captioning model is illustrated in \autoref{fig:architecture}. The model consists of dual vision encoders, a Query Transformer (Q-Former), and a domain-adapted LLaMA decoder, which are described in detail in the following subsections.

\subsubsection{Dual Encoder}
\label{sec:dual-encoder}

To derive robust and semantically rich visual representations from medical images, we adopt an ensemble of two vision encoders. Specifically, we utilize SigLIP2 \cite{11}, a general-purpose image encoder pretrained on large-scale natural image-text pairs, and BioMedCLIP \cite{12}, a medical-domain-specific encoder trained on 15 million image-caption pairs mined from PubMed Central. To address the lack of medical image knowledge in the original SigLIP2 model, we perform domain-specific pre-adaptation by fine-tuning it on the dataset provided by the ImageCLEF2025 Caption Prediction Task \cite{28,29}. This improves the encoder’s ability to capture domain-relevant visual features while preserving generalization capacity.

Following standard practice, we remove the classification heads from both encoders and use the token-level hidden states as feature maps. Let the output token sequences of BioMedCLIP and SigLIP2 be $\mathbf{H}_{\text{bioclip}} \in \mathbb{R}^{B \times T_{\text{bio}} \times d_{\text{bio}}}$ and $\mathbf{H}_{\text{siglip}} \in \mathbb{R}^{B \times T_{\text{sig}} \times d_{\text{sig}}}$, respectively. We then compare three strategies for fusing these two sequences before passing them to the Q-Former as encoder hidden states.

First, in the simple feature concatenation baseline, we aim to minimally process the encoder outputs before feeding them into the Q-Former. Concretely, we extract a global representation from each sequence, denoted $\mathbf{f}_{\text{bioclip}} \in \mathbb{R}^{B \times d_{\text{bio}}}$ and $\mathbf{f}_{\text{siglip}} \in \mathbb{R}^{B \times d_{\text{sig}}}$, for example by applying global average pooling or taking a special classification token. These vectors are concatenated along the feature dimension to obtain $\mathbf{f} = [\mathbf{f}_{\text{bioclip}} ; \mathbf{f}_{\text{siglip}}] \in \mathbb{R}^{B \times D}$, where $D = d_{\text{bio}} + d_{\text{sig}}$. The resulting $\mathbf{f}$ is treated as a length-1 token sequence and used as the encoder hidden state for Q-Former cross-attention. This design corresponds to a late-fusion scheme that preserves the pretrained encoders' representations with minimal distortion.

Second, in Bi-Directional Self-Attention Fusion, we concatenate the token sequences from the two encoders along the sequence dimension and feed the result into a lightweight multi-layer Transformer encoder to obtain a jointly contextualized representation through global self-attention. The fused token sequence takes the form $\mathbf{H}_{\text{fuse}} \in \mathbb{R}^{B \times (T_{\text{bio}} + T_{\text{sig}}) \times d}$, where $d$ denotes the hidden dimension of the fusion module, and $\mathbf{H}_{\text{fuse}}$ is directly used as the encoder hidden state for the Q-Former.

Third, in Dual Cross-Attention Fusion, we keep the SigLIP2 and BioMedCLIP token sequences as two separate streams and apply a stack of bi-directional cross-attention blocks so that the two representations explicitly exchange information. The outputs of the final block are then concatenated to form a fused token sequence, which is passed to the Q-Former as the encoder hidden state.

All three fusion schemes are implemented within an otherwise identical captioning architecture and training configuration, allowing analysis of fusion behavior independent of decoder or optimization changes.

\subsubsection{Query Transformer (Q-Former)}
To reduce redundancy and computational burden, we apply a Q-Former~\cite{14} that compresses the encoder hidden states into a fixed set of informative latent tokens. Let $\mathbf{H}_{\text{enc}} \in \mathbb{R}^{B \times T_e \times d_e}$ denote the encoder hidden states provided by the selected fusion scheme. Let $\mathbf{Q}_0 \in \mathbb{R}^{B \times N_q \times d_q}$ denote a learnable set of query tokens. The Q-Former maps $(\mathbf{Q}_0, \mathbf{H}_{\text{enc}})$ to $\mathbf{Z} \in \mathbb{R}^{B \times N_q \times d_q}$ through stacked Transformer layers composed of self-attention over the query tokens and cross-attention to $\mathbf{H}_{\text{enc}}$. This output is used for both caption generation and auxiliary concept classification.

To enhance medical grounding, we incorporate a multitask classification objective~\cite{27}. The output $\mathbf{Z}$ is mean-pooled across the query dimension to produce a global representation $\bar{\mathbf{z}} \in \mathbb{R}^{B \times d_q}$. This representation is passed through two linear classifiers: one to predict concept presence among $C$ Concept Unique Identifiers (CUIs) and another to predict $T$ coarse concept types. The overall training objective is
\begin{equation}
\mathcal{L}_{\text{total}}
=
\mathcal{L}_{\text{caption}}
+
\lambda \mathcal{L}_{\text{cls}},
\end{equation}
where $\mathcal{L}_{\text{caption}}$ denotes the cross-entropy loss over caption tokens, $\mathcal{L}_{\text{cls}}$ denotes the auxiliary multi-label margin loss, and $\lambda$ is the weighting factor for the auxiliary term.

\subsubsection{Caption Decoder}
For the caption generation task, we adopt Bio-Medical LLaMA-3-8B \cite{15}, a domain-specialized variant of Meta-Llama-3-8B-Instruct \cite{30}, as the language decoder. The model has been fine-tuned on BioMedData, a high-quality biomedical dataset containing over 500,000 entries. The dataset comprises a blend of synthetic and manually curated samples, enabling robust generalization across a wide range of biomedical contexts. During training, the Q-Former output tokens are inserted as prefix embeddings that condition each decoding step on visual evidence. To enable efficient fine-tuning, we incorporate LoRA \cite{16} modules into the decoder. This allows the model to adapt to medical image captioning tasks with minimal parameter updates while preserving the core language modeling capabilities of LLaMA.

\subsection{Post-processing}
To further refine the raw captions generated by six distinct captioning models, we applied post-processing strategies to improve both clinical consistency and factual relevance. This section presents two major post-processing components: (1) summarization-based refinement using GPT APIs and (2) candidate caption reranking based on semantic and domain-specific metrics.

\subsubsection{Summarization-based Refinement}
We employed two GPT-4-based strategies for summarization \cite{17} to consolidate the six candidate captions—each produced by a different model—into a single, medically accurate sentence. Both approaches aimed to improve readability, reduce redundancy, and ensure consistency with structured medical knowledge. The exact prompts used for each summarization method are provided in \autoref{tab:prompt_templates}.

\begin{table}
\caption{Prompt templates used for GPT-4-based medical image caption summarization.}
\label{tab:prompt_templates}
\setlength{\tabcolsep}{3pt}
\begin{tabular}{|p{118pt}|p{118pt}|}
\hline
\textbf{Chain-of-Thought Summarization Template} &
\textbf{Prompt-guided Summarization Template} \\
\hline

You are a board-certified radiologist. \par
\textbf{Task:} \par
1) Parse each caption and list semantic components such as modality, anatomic site, and pathologies. \par
2) Build a consensus table based on token frequency. \par
3) Resolve conflicts by majority vote or by selecting the more specific description. \par
4) Compose one radiology-style sentence (approximately 35--45 words). \par
\textbf{Output:} FINAL\_CAPTION: \textless your summary\textgreater \par
\textbf{Captions:} \par
caption 1: \{caption1\} \par
caption 2: \{caption2\} \par
caption 3: \{caption3\} \par
caption 4: \{caption4\} \par
caption 5: \{caption5\} \par
caption 6: \{caption6\}

&

You are a radiologist summarizing multiple captions of a medical image into one detailed sentence. \par
\textbf{Instructions:} \par
1) Integrate the imaging modality, anatomical location, pathological findings, and relevant clinical details. \par
2) Use medically correct and extractive phrasing with high token overlap; avoid paraphrasing unless synonymous medical terminology improves clarity. \par
3) Use present continuous tense with a subject--predicate--object structure. \par
4) Keep the summary natural, clinically accurate, and around 40 words. \par
5) If captions are inconsistent, prioritize findings with the highest diagnostic or therapeutic relevance. \par
\textbf{Captions:} \par
caption 1: \{caption1\} \par
caption 2: \{caption2\} \par
caption 3: \{caption3\} \par
caption 4: \{caption4\} \par
caption 5: \{caption5\} \par
caption 6: \{caption6\} \\
\hline

\multicolumn{2}{p{242pt}}{The Chain-of-Thought strategy decomposes candidate captions into semantic units and aggregates them through consensus, whereas the Prompt-guided strategy directly synthesizes multiple captions into a clinically coherent single-sentence summary. Both templates are designed to generate standardized radiology-style descriptions from diverse candidate captions.}
\end{tabular}
\end{table}

\emph{Prompt-guided summarization:} Six caption candidates, one produced by each captioning model, were aggregated and provided to GPT-4 using a fixed prompt template. The prompt requested a concise and clinically coherent summary under the assumption that all captions corresponded to the same medical image. This helped reduce redundancy, resolve inconsistencies, and unify expression styles across captions.

\vspace{0.5\baselineskip}

\emph{Chain-of-Thought Summarization:} In this variant, the prompt explicitly encouraged chain-of-thought reasoning \cite{32} before concluding the final summary. The intent was to increase factual consistency by encouraging the model to align each summary point with the underlying clinical evidence extracted from input captions. Empirically, this strategy improved alignment with structured medical entities.

\subsubsection{Caption Reranking}
\begin{figure}[ht]
    \centering
    \includegraphics[width=0.9\linewidth]{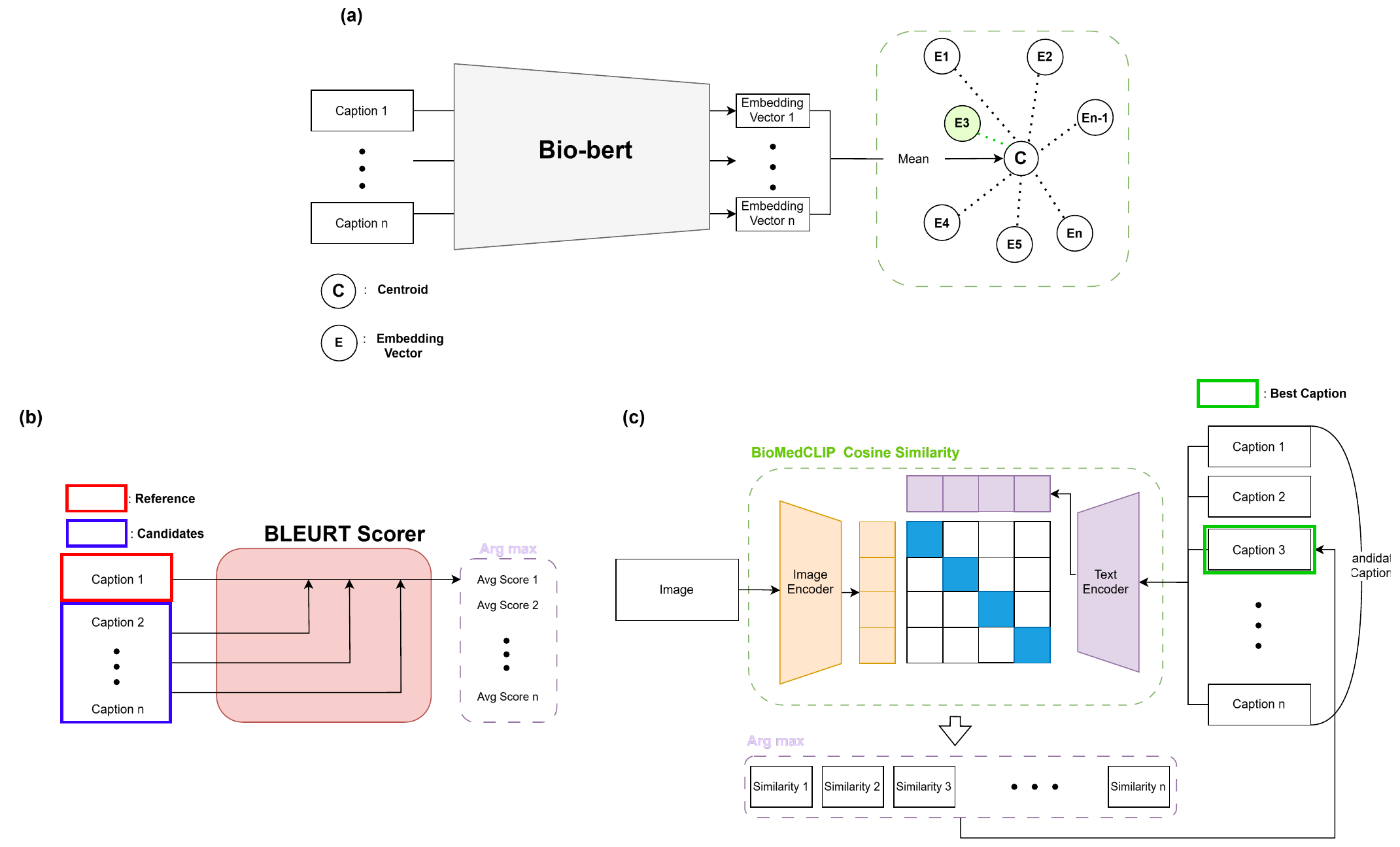}
    \caption{Overview of reranking methods: (a) BioBERT centroid similarity, (b) BLEURT self-consensus, (c) BioMedCLIP image-text alignment}
    \label{fig:reranking}
\end{figure}

To select the most appropriate caption among the generated candidates, we implemented a reranking module based on three different metrics: BioMedCLIP image--text alignment, BLEURT self-consensus, and BioBERT centroid proximity. The overall framework of these reranking strategies is illustrated in \autoref{fig:reranking}.

\emph{BioMedCLIP image--text alignment:}
Each caption $c_i$ is embedded into $\mathbf{v}_i$ using the BioMedCLIP \cite{12} text encoder, following the BioMedCLIP-based reranking strategy in \cite{13}, and the corresponding image is embedded into $\mathbf{w}$ using the image encoder. We compute the cosine similarity
\[
\mathrm{sim}_i = \cos(\mathbf{v}_i, \mathbf{w}) = \frac{\mathbf{v}_i \cdot \mathbf{w}}{\lVert \mathbf{v}_i \rVert \cdot \lVert \mathbf{w} \rVert},
\]
which measures image--text alignment in a biomedical semantic space. The selected caption is $\hat{c} = \arg\max_i \mathrm{sim}_i$.

\vspace{0.5\baselineskip}

\emph{BLEURT self-consensus:}
BLEURT \cite{18} estimates sentence quality via a regression head over BERT-style embeddings. For caption $c_i$ among $n$ candidates, we compute the leave-one-out average
\[
\mathrm{score}_i = \frac{1}{n - 1} \sum_{j \ne i} \mathrm{BLEURT}(c_i, c_j),
\]
which favors captions that are semantically central to the candidate set and therefore robust to outliers. The selected caption is $\hat{c} = \arg\max_i \mathrm{score}_i$.

\vspace{0.5\baselineskip}

\emph{BioBERT centroid proximity:}
All captions are embedded via BioBERT \cite{20} as vectors $\mathbf{v}_1, \ldots, \mathbf{v}_n$. The centroid
\[
\mathbf{v}_c = \frac{1}{n} \sum_{i=1}^{n} \mathbf{v}_i
\]
represents the consensus semantic position. Each caption is then ranked by its Euclidean distance to the centroid \cite{19}, and the selected caption is $\hat{c} = \arg\min_i \lVert \mathbf{v}_i - \mathbf{v}_c \rVert$.

\subsection{Reinforcement Learning with MedPAIR-SCST}
\label{sec:medpair-scst}

To further align the decoder with clinically meaningful semantics, we fine-tune the model using a self-critical sequence training (SCST) objective augmented with a clinically informed reward. For each image $x_b$, the current decoder policy samples $K$ candidate captions, $\{c_{b,1}, \dots, c_{b,K}\}$, using temperature-scaled top-$k$ sampling and, optionally, top-$p$ sampling.

Given the ground-truth caption $y_b$, each candidate is assigned a composite reward,
\begin{equation}
\begin{aligned}
R(c,y) = \frac{1}{3}\Big(&
\text{BERTScore}_{\text{F1}}(c', y') \\
&+ \text{ROUGE-1}_{\text{F1}}(c', y') \\
&+ \text{UMLS-F1}(c, y)
\Big).
\end{aligned}
\end{equation}
where $c'$ and $y'$ denote lightly normalized captions obtained by lowercasing and replacing digits with the token ``number,'' whereas $c$ and $y$ denote the corresponding raw texts. BERTScore is computed with IDF reweighting, ROUGE-1 is measured as unigram F1, and UMLS-F1 is defined as the F1 score between the sets of Concept Unique Identifiers (CUIs) extracted from the raw texts using MedCAT. This reward jointly encourages lexical and semantic similarity as well as accurate recovery of UMLS concepts.

To obtain sequence-level log-likelihoods for policy-gradient updates, all sampled captions are re-scored in a single teacher-forcing pass. Specifically, the image-conditioned prefix representation is concatenated with the token embeddings of each sampled caption, and the decoder is run in teacher-forcing mode. For each caption $c_{b,k}$, we sum token log-probabilities up to the first EOS token, mask padding positions, and normalize by the effective sequence length:
\begin{equation}
\bar{\ell}_{b,k}
= \frac{1}{T_{b,k}} \sum_{t=1}^{T_{b,k}}
\log p_\theta\big(c_{b,k,t} \mid c_{b,k,<t}, x_b\big),
\end{equation}
where $T_{b,k}$ denotes the number of valid tokens before EOS. We use $\bar{\ell}_{b,k}$ as a length-normalized sequence score to reduce trivial length effects.

Within each image-specific group $b$, the sampled rewards $\{R_{b,1}, \dots, R_{b,K}\}$ are first centered and then converted into normalized weights through a softmax with temperature $\tau$:
\begin{equation}
w_{b,k} = \frac{\exp\big((R_{b,k} - \bar{R}_b)/\tau\big)}
{\sum_{j=1}^{K} \exp\big((R_{b,j} - \bar{R}_b)/\tau\big)},
\end{equation}
where $\bar{R}_b$ is the mean reward for image $b$. We then define group-wise advantages as
\begin{equation}
a_{b,k} = w_{b,k} - \frac{1}{K},
\end{equation}
so that $\sum_{k=1}^{K} a_{b,k} = 0$ and only relative reward differences within the group contribute to the update. The resulting group-wise SCST loss is
\begin{equation}
\mathcal{L}_{\text{group}}
= - \frac{1}{B} \sum_{b=1}^{B} \sum_{k=1}^{K}
a_{b,k} \, \bar{\ell}_{b,k},
\end{equation}
which increases the likelihood of captions with positive group-wise advantages and decreases that of captions with negative advantages.

To align model scores more explicitly with the clinically informed reward, we further introduce a reference-free pairwise ranking loss. For each image $b$, we consider all ordered index pairs
\(
\mathcal{P}_b = \{(i,j) \mid i \neq j,\; R_{b,i} > R_{b,j}\}
\),
and penalize inconsistencies between reward ordering and model scores using a softplus margin:
\begin{equation}
\mathcal{L}_{\text{pair}}
= \frac{1}{\sum_b |\mathcal{P}_b|}
\sum_{b=1}^{B} \sum_{(i,j) \in \mathcal{P}_b}
w_{b,ij} \, \mathrm{softplus}\!\Big(
m - \big(\bar{\ell}_{b,i} - \bar{\ell}_{b,j}\big)
\Big),
\end{equation}
where $m$ is a small margin and $w_{b,ij}$ is a non-negative weight that increases with the reward gap $R_{b,i} - R_{b,j}$. This loss can be interpreted as a logistic pairwise preference objective defined on differences in sequence log-likelihood, encouraging higher-reward captions to receive higher model scores without requiring a separate frozen reference policy or KL regularization.

The main reinforcement learning objective is defined as
\begin{equation}
\mathcal{L}_{\text{MedPAIR}}
= \mathcal{L}_{\text{group}} + \lambda_{\text{pair}} \, \mathcal{L}_{\text{pair}},
\end{equation}
where $\lambda_{\text{pair}}$ controls the strength of the ranking signal. In practice, we additionally apply small regularization terms to the sampling weights and token-level likelihoods, such as entropy-based penalties and a mild fluency constraint, to improve optimization stability. For clarity, these auxiliary stabilizers are omitted from the main objective.

In summary, the proposed \emph{Medical Pairwise Aligned Image-captioning with Reinforcement SCST} (MedPAIR-SCST) objective combines (i) SCST-style self-critical optimization with group-normalized advantages, (ii) a clinically informed reward that explicitly scores UMLS concept recovery in addition to lexical and semantic similarity, and (iii) a reference-free pairwise ranking term that aligns reward ordering with model likelihoods. Together with lightweight auxiliary regularizers for optimization stability, this design improves both linguistic quality and clinical factuality without maintaining a separate frozen reference model, thereby avoiding the additional memory and system complexity typically associated with KL-regularized reinforcement learning fine-tuning.

\section{Experiments}
\subsection{Experimental Setups}

We conduct experiments on the extended version of the ROCOv2 dataset \cite{28, 33}, specifically curated for the ImageCLEFmedical 2025 Caption Prediction Task \cite{28}. Unlike the original ROCOv2 \cite{33}, this updated release includes additional manual annotations as well as a newly introduced test set for the 2025 challenge. The dataset configuration differs from prior versions: the previous test set from ROCOv2 has been reassigned as the validation set, and the prior validation set has been merged into the training set. The newly collected 2025 test set contains unseen images to evaluate generalization performance under updated task conditions. However, because the ground-truth captions for the 2025 test set are not released even after the end of the challenge, we report all quantitative results on the validation split. The resulting splits comprise 80,091 training images and 17,277 validation images. Each image is associated with a manually curated caption and UMLS concepts, making it suitable for both generation and concept detection tasks.

Model performance is evaluated using four metrics that assess both relevance and factuality. Relevance is measured with BERTScore (Recall with IDF), ROUGE-1 (F1), and BLEURT. BERTScore is computed with the microsoft/deberta-xlarge-mnli model using IDF scores derived from the validation set, and BLEURT uses the recommended BLEURT-20 checkpoint. All relevance metrics are calculated on lowercase, punctuation-free captions with numbers replaced by the token “number.” For factuality, UMLS Concept F1 is computed using MedCAT\cite{35} with semantic type filtering via QuickUMLS. 

Our model employs either BioMedCLIP or SigLIP2 as standalone vision encoders, or an ensemble configuration obtained by channel-wise concatenation of their outputs. When using Bio-Medical LLaMA-3-8B as the language decoder, visual features are processed by a 6-layer Q-Former with 32 learnable query tokens. In this configuration, the Q-Former maps the concatenated 2{,}304-dimensional visual representation to 4{,}096-dimensional embeddings compatible with the 8B decoder. The model optionally includes auxiliary classification heads that jointly predict 2{,}478 UMLS concepts and 21 coarse semantic types. The overall training objective is defined as a weighted sum of the captioning loss and the concept classification loss, with the weighting factor $\lambda$ fixed to 0.3. Training is performed with the AdamW optimizer; the learning rate is linearly increased to $1\times 10^{-4}$ during the first epoch and then annealed to $1\times 10^{-6}$ over a total of 10 epochs. Experiments with the 8B decoder are conducted on a single NVIDIA H100 GPU with a batch size of 16 and a gradient accumulation step of~2. At inference time, we use beam search with a beam width of 3, a repetition penalty of 2.5, a length penalty of 2.0, and a minimum and maximum generation length of 8 and 64 tokens, respectively.

In the lightweight 1B configuration, we keep the same vision encoder choices and Q-Former design (6 layers and 32 learnable query tokens), but project a 1{,}536-dimensional concatenated visual representation into a 2{,}048-dimensional embedding space to match the hidden size of the 1B decoder. The auxiliary heads and loss formulation (including $\lambda = 0.3$) are identical to those of the 8B model. Experiments with the 1B decoder are run on a single NVIDIA A100 GPU, while all remaining training and decoding hyperparameters (optimizer, number of epochs, learning rate schedule, batch size, gradient accumulation, and beam search configuration) are kept the same as in the 8B setting.

Additionally, for the 1B decoder, we perform a second-stage fine-tuning experiment, denoted MedPAIR-SCST, on top of the dual-encoder + auxiliary-task model after completing the first-stage supervised training. In this phase, we initialize from the stage-1 checkpoint and apply Self-Critical Sequence Training (SCST) for 2 epochs, evaluating on the validation set every 0.1 epoch to closely track performance. During SCST, we sample $K = 4$ candidate captions per image with a maximum generation length of 40 tokens. Sampling hyperparameters are set to a temperature of 0.9, top-$k = 40$, and top-$p = 0.85$; the pairwise ranking component uses a margin of 0.02 and a pairwise weight of 0.3. Optimization is carried out with AdamW and a cosine-style learning rate schedule bounded between a minimum of $5\times 10^{-6}$ and a maximum of $1\times 10^{-5}$, while all other optimization settings follow those of the supervised stage.

\subsection{Base Model}

\begin{table}[t]
\centering
\footnotesize
\caption{Summarizes the performance of six backbone configurations that differ in (i) vision-encoder composition and (ii) the presence of auxiliary concept heads. All models share the Q-Former and a BioMedLLaMA-3-8B decoding stack and are trained under the identical hyper-parameter schedule.}
\label{tab:8b_results}
\setlength{\tabcolsep}{3pt}
\renewcommand{\arraystretch}{1.1}
\begin{tabular}{|p{1.6cm}|p{0.6cm}|p{1.0cm}|p{1.2cm}|p{1.1cm}|p{1.0cm}|}
\hline
Encoder & Aux. & BERT-R & ROUGE-1 & BLEURT & UMLS F1 \\
\hline
BioMedCLIP   & X & 0.5845 & 0.2261 & 0.3100 & 0.1405 \\
SigLIP2      & X & 0.5796 & 0.2194 & 0.3047 & 0.1397 \\
Dual Encoder & X & 0.5826 & 0.2305 & 0.3133 & 0.1514 \\
BioMedCLIP   & O & 0.5860 & 0.2252 & 0.3100 & 0.1411 \\
SigLIP2      & O & 0.5837 & 0.2289 & 0.3090 & 0.1438 \\
Dual Encoder & O & \textbf{0.5863} & \textbf{0.2347} & \textbf{0.3150} & \textbf{0.1528} \\
\hline
\end{tabular}
\end{table}

\noindent 
\autoref{tab:8b_results} reports the results obtained with the 8B decoder as we vary the choice of vision encoder and the presence of an auxiliary UMLS concept classification head. Among single-encoder configurations without the auxiliary task, BioMedCLIP outperforms SigLIP2 in both BERTScore Recall (0.5845 vs.\ 0.5796) and BLEURT (0.3100 vs.\ 0.3047), while the dual-encoder model further improves BLEURT to 0.3133 and achieves the highest UMLS F1 (0.1514) within this group. Adding the UMLS concept and type prediction auxiliary head consistently improves factuality while preserving or slightly enhancing relevance across all vision encoders: UMLS F1 increases from 0.1405 to 0.1411 for BioMedCLIP and from 0.1397 to 0.1438 for SigLIP2. The dual-encoder with the auxiliary task attains the best overall performance, achieving the highest scores on all four metrics (BERTScore Recall 0.5863, ROUGE-1 0.2347, BLEURT 0.3150, and UMLS F1 0.1528), indicating that combining complementary visual features with an explicit UMLS concept classification head is most effective for jointly improving semantic relevance and clinical faithfulness in the 8B decoder setting.

\begin{table}[t]
\centering
\footnotesize
\caption{Summarizes the performance of six backbone configurations that differ in (i) vision-encoder composition and (ii) the presence of auxiliary concept heads. All models share the Q-Former and a BioMedLLaMA-3-1B decoding stack and are trained under the identical hyper-parameter schedule.}
\label{tab:1b_results}
\setlength{\tabcolsep}{3pt}
\renewcommand{\arraystretch}{1.1}
\begin{tabular}{|p{1.6cm}|p{0.6cm}|p{1.0cm}|p{1.0cm}|p{1.2cm}|p{1.1cm}|p{1.0cm}|}
\hline
Encoder & Aux. & BERT-R & BERT-R w/o IDF & ROUGE-1 & BLEURT & UMLS F1 \\
\hline
BioMedCLIP   & X & 0.5714 & 0.6284 & 0.2285 & 0.3025 & 0.1317 \\
SigLIP2      & X & 0.5632 & 0.6227 & 0.2183 & 0.2937 & 0.1197 \\
Dual Encoder & X & \textbf{0.5775} & 0.6246 & \textbf{0.2382} & \textbf{0.3098} & 0.1450 \\
BioMedCLIP   & O & 0.5625 & 0.6189 & 0.2167 & 0.3000 & 0.1229 \\
SigLIP2      & O & 0.5577 & 0.6157 & 0.2067 & 0.2921 & 0.1099 \\
Dual Encoder & O & 0.5734 & \textbf{0.6298} & 0.2334 & 0.3065 & \textbf{0.1463} \\
\hline
\end{tabular}
\end{table}

\autoref{tab:1b_results} reports the results obtained with the 1B decoder as we vary the choice of vision encoder and the presence of an auxiliary UMLS concept classification head. Among configurations without the auxiliary task, the dual-encoder model clearly outperforms the single-encoder variants, achieving the highest BERTScore Recall (0.5775), ROUGE-1 (0.2382), BLEURT (0.3098), and UMLS F1 (0.1450). In contrast, adding the auxiliary UMLS head to the single-encoder models yields no benefit and even degrades performance: for BioMedCLIP, UMLS F1 drops from 0.1317 to 0.1229, and for SigLIP2 from 0.1197 to 0.1099, accompanied by small decreases in relevance metrics. For the dual-encoder, the auxiliary task leads to a modest improvement in UMLS F1 (from 0.1450 to 0.1463) and BERTScore (Recall, no IDF; from 0.6246 to 0.6298), while slightly reducing BERTScore Recall, ROUGE-1, and BLEURT. Overall, in the 1B decoder setting the dual-encoder architecture remains effective, but the gains from introducing the UMLS concept classification head are limited and substantially weaker than those observed with the 8B decoder.

\subsection{Fusion Method}

\begin{table}[t]
\centering
\footnotesize
\caption{Comparison of three dual-encoder fusion strategies in terms of captioning and clinical concept metrics.}
\label{tab:fusion_strategies}
\setlength{\tabcolsep}{3pt}
\renewcommand{\arraystretch}{1.1}
\begin{tabular}{|p{2.0cm}|p{1.0cm}|p{1.2cm}|p{1.2cm}|p{1.1cm}|p{1.0cm}|}
\hline
Fusion method & BERT-R & BERT-R w/o IDF & ROUGE-1 & BLEURT & UMLS F1 \\
\hline
Simple Concat & 0.5734 & 0.6298 & 0.2334 & 0.3065 & 0.1463 \\
SA Fusion & 0.5383 & 0.5394 & 0.1839 & 0.2893 & 0.1096 \\
CA Fusion & 0.5570 & - & 0.2064 & 0.2893 & 0.1238 \\
\hline
\end{tabular}
\end{table}

In Section~\ref{sec:dual-encoder}, we introduced three strategies for combining the outputs of the two visual encoders: simple feature concatenation, Bi-Directional Self-Attention Fusion, and Dual Cross-Attention (CA) Fusion. In this subsection, we compare these alternatives under an otherwise identical captioning architecture and training setup. The quantitative results are summarized in \autoref{tab:fusion_strategies}, which reports BERTScore-Recall (with and without IDF), ROUGE-1, BLEURT, and UMLS concept F1.

Across all reported metrics, the simple concatenation baseline achieves the best overall performance. It attains the highest BERTScore-Recall (0.5734), ROUGE-1 (0.2334), BLEURT (0.3065), and UMLS F1 (0.1463). Both attention-based fusion variants perform worse. Bi-Directional Self-Attention Fusion shows the largest drop, with BERTScore-Recall decreasing to 0.5383, ROUGE-1 to 0.1839, BLEURT to 0.2893, and UMLS F1 to 0.1096. Dual Cross-Attention Fusion shows a more moderate decline relative to simple concatenation (BERTScore-Recall 0.5570, ROUGE-1 0.2064, BLEURT 0.2893, and UMLS F1 0.1238), but it still does not outperform the concatenation baseline.

These results suggest that, under the limited medical image--report data regime considered here, adding extra self-attention or cross-attention fusion blocks to more tightly couple the dual-encoder representations does not improve downstream performance and may instead make optimization less effective. In contrast, preserving the representations produced by each pretrained encoder and allowing the Q-Former to attend directly over the concatenated token sequence leads to stronger and more consistent results across both captioning and clinical concept metrics. This trend is broadly consistent with prior observations that simple late-fusion strategies can remain competitive in multimodal settings, including UniCat~\cite{55}, where simple late-fusion strategies based on independently pretrained encoders and embedding concatenation were shown to be robust and effective. Taken together, our results suggest that, in data-constrained medical imaging settings, it is often preferable to preserve the strengths of each pretrained encoder and adopt minimal fusion, rather than introducing more complex fusion modules.

\subsection{Post-Processing}

\begin{table}[t]
\centering
\footnotesize
\caption{Validation-only comparison of three reranking strategies applied to six candidate captions per image generated by the 8B decoder. The best score for each metric is shown in bold. (* candidates include GPT-4 Chain-of-Thought and prompt-guided summaries.)}
\label{tab:reranking_results}
\setlength{\tabcolsep}{3pt}
\renewcommand{\arraystretch}{1.1}
\begin{tabular}{|p{2.2cm}|p{1.2cm}|p{1.2cm}|p{1.2cm}|p{1.2cm}|}
\hline
Reranker & BERT-R & ROUGE-1 & BLEURT & UMLS F1 \\
\hline
BioMedCLIP      & 0.5873 & 0.2338 & 0.3130 & 0.1499 \\
BLEURT*         & 0.5880 & 0.2368 & 0.3178 & 0.1539 \\
bioBERT         & \textbf{0.5922} & 0.2409 & \textbf{0.3179} & \textbf{0.1552} \\
Base Model (8B) & 0.5826 & \textbf{0.2440} & 0.3176 & 0.1547 \\
\hline
\end{tabular}
\end{table}

As summarized in \autoref{tab:reranking_results}, reranking based on a single embedding-based score criterion consistently shifts the performance profile across metrics. The bioBERT selector achieves the best BERTScore Recall (0.5922), BLEURT (0.3179), and UMLS F1 (0.1552), while the BLEURT-based selector is a close second (0.5880 / 0.3178 / 0.1539). By comparison, the BioMedCLIP selector performs worse on all reported metrics (BERTScore Recall 0.5873, ROUGE-1 0.2338, BLEURT 0.3130, UMLS F1 0.1499). Notably, the non-reranked base model retains the highest ROUGE-1 (0.2440), suggesting that inference-time candidate selection tends to improve semantic and clinical alignment (BERTScore, BLEURT, and UMLS F1) at the expense of a small reduction in n-gram overlap.

In addition to reranking, we also evaluated GPT-4-guided summarization as an alternative post-processing strategy. Although aggregating multiple candidate captions into a single concise output is intuitively appealing, the summarization-based approaches performed worse than reranking in our quantitative evaluations. In particular, both prompt-guided summarization and chain-of-thought prompting often produced less precise outputs and occasionally introduced clinically irrelevant or hallucinated content. These weaknesses were especially apparent in factual grounding metrics such as UMLS F1, suggesting that generative summarization may abstract away or omit clinically important concepts during compression. A more detailed comparison of these behaviors is provided in \autoref{tab:appen}.

\subsection{Comparison with other models}

\begin{table}[t]
\centering
\footnotesize
\caption{Comparison of our 1B base model and MedPAIR-SCST against prior methods across text quality and clinical concept fidelity metrics on the medical image captioning benchmark.}
\label{tab:other}
\setlength{\tabcolsep}{3pt}
\renewcommand{\arraystretch}{1.1}
\begin{tabular}{|p{2.4cm}|p{1.0cm}|p{1.2cm}|p{1.0cm}|p{1.0cm}|p{1.0cm}|}
\hline
Method & BERT-R & BERT-R w/o IDF & ROUGE-1 & BLEURT & UMLS F1 \\
\hline
R2Gen \cite{56} & 0.5519 & 0.5967 & 0.2212 & 0.2710 & 0.1495 \\
CvTdistilGPT2 \cite{57} & 0.5805 & 0.6207 & 0.2299 & 0.2867 & 0.1318 \\
Base Model(1B) & 0.5775 & 0.6246 & 0.2382 & 0.3098 & 0.1450 \\
Base Model(1B) + MedPAIR-SCST & 0.6000 & 0.6584 & 0.2755 & 0.3122 & 0.1821 \\
\hline
\end{tabular}
\end{table}

We evaluated the effectiveness of our approach by comparing it with established medical image captioning methods. The quantitative results are reported in \autoref{tab:other}. Our 1B base model shows competitive, and in some cases improved, performance relative to R2Gen~\cite{56} and CvTdistilGPT2~\cite{57}. In particular, the base model achieves higher ROUGE-1 (0.2382) and BLEURT (0.3098), indicating better surface-level alignment and overall generation quality.

Further improvements are observed when MedPAIR-SCST is applied to the base model, as shown in \autoref{tab:other}. BERTScore Recall increases from 0.5775 to 0.6000, and BERTScore Recall without IDF rises from 0.6246 to 0.6584. ROUGE-1 shows a substantial gain from 0.2382 to 0.2755, while BLEURT improves slightly from 0.3098 to 0.3122. Most notably, UMLS F1 increases from 0.1450 to 0.1821, suggesting that MedPAIR-SCST improves clinically grounded concept fidelity in addition to linguistic quality. Overall, these results support the effectiveness of our reinforcement-learning-based optimization in producing more clinically faithful captions.

\section{Conclusion}
This study treats clinical alignment as the central objective of medical image captioning and proposes a two-axis alignment framework that separates training-time alignment from inference-time alignment, allowing their effects to be examined independently. We build a baseline captioning model consisting of dual visual encoders (BioMedCLIP and SigLIP2), a Q-Former with learnable query tokens, and a LLaMA-based decoder, and we systematically analyze, under matched training conditions, the contributions of (i) vision-encoder composition and (ii) auxiliary UMLS concept/type classification heads. Under the 8B decoder setting, the configuration combining dual encoders with auxiliary UMLS supervision achieves the most stable and strongest performance across BERTScore, ROUGE-1, BLEURT, and UMLS F1, suggesting that complementary visual representations and explicit concept learning can jointly improve semantic relevance and clinical factuality. In contrast, under the 1B decoder setting, although the benefits of dual encoders largely remain, the auxiliary heads provide limited gains and can even degrade performance, indicating that the effectiveness of auxiliary supervision depends on decoder capacity and the model’s ability to absorb additional training signals.

For inference-time alignment, we apply embedding-based reranking over candidates generated by six independently trained models. Selection strategies based on BioBERT centroid proximity and BLEURT self-consensus consistently improve BERTScore, BLEURT, and UMLS F1, demonstrating that even lightweight post-hoc selection can substantially enhance clinical relevance and concept fidelity. By contrast, GPT-4-based summarization refinement, despite its intuitive appeal, underperforms reranking in quantitative evaluation, plausibly because compressive rewriting may dilute salient findings or introduce extraneous content. Moreover, in the comparison of dual-encoder fusion designs, we find that simple token concatenation yields better overall results across metrics than more complex self-attention or cross-attention fusion modules, reinforcing the effectiveness of minimal fusion in data-constrained clinical settings.

For training-time alignment, we propose MedPAIR-SCST, a reinforcement-style objective that operates stably without a reference model or KL regularization. By combining group-level reward aggregation with a reference-free pairwise ranking term, MedPAIR-SCST shifts the model’s generation distribution toward clinically more accurate outputs and yields substantial gains, particularly in UMLS F1, while preserving linguistic quality. Comparisons with prior baselines such as R2Gen and CvTdistilGPT2 show that our 1B-based model is competitive, and that applying MedPAIR-SCST further improves alignment, supporting the practical value of distribution-level optimization for clinical faithfulness.

Nevertheless, this work has several limitations. First, because the ground-truth annotations for the ImageCLEFmedical 2025 test split are not publicly available, our quantitative analyses rely on the validation split, which limits our ability to fully assess generalization to hidden test data. Second, our reward design is based on a combination of BERTScore, ROUGE-1, and UMLS F1, which may still introduce biases toward particular surface forms or terminology distributions. Third, the limited effectiveness of auxiliary UMLS supervision in the 1B setting suggests that interactions among auxiliary task difficulty, loss weighting, label quality, and decoder capacity require more careful study.

Future directions include: (1) evaluation under hidden-test conditions and on external multi-institutional datasets to better quantify clinical generalization; (2) the design of more structured factuality rewards that encode modality-, anatomy-, and finding-level constraints while explicitly modeling boundary cases and contradictions between concepts; (3) improved auxiliary learning efficiency for lightweight decoders through loss-weight scheduling and hierarchical target formulations; and (4) extension of MedPAIR-SCST to larger decoders, together with minimal-fusion principles, to achieve stronger clinical reliability under limited-data regimes.

In conclusion, this study provides a systematic treatment of selection-based inference alignment and distribution-based training alignment for medical image captioning. Through principled dual-encoder feature composition and the reference-free MedPAIR-SCST objective, we advance clinically more reliable caption generation and provide empirical evidence that, in resource- and data-constrained medical imaging settings, carefully designed alignment signals and their placement can be more effective than increasing architectural complexity.

\appendix[Summarization-Based Refinement Results]

\begin{table}[t]
\centering
\footnotesize
\caption{Quantitative comparison of GPT-based summarization methods for medical caption refinement using candidate captions generated by the 8B decoder.}
\label{tab:appen}
\setlength{\tabcolsep}{3pt}
\renewcommand{\arraystretch}{1.1}
\begin{tabular}{|p{2.0cm}|p{1.2cm}|p{1.2cm}|p{1.2cm}|p{1.2cm}|}
\hline
Method & BERT-R & ROUGE-1 & BLEURT & UMLS F1 \\
\hline
CoT & 0.5705 & 0.2032 & 0.3197 & 0.1236 \\
Prompt-guided & 0.5723 & 0.2016 & 0.3176 & 0.1242 \\
\hline
\end{tabular}
\end{table}

GPT-4-based summarization of candidate captions can produce more concise outputs; however, it did not lead to consistent improvements in clinical alignment. For instance, the prompt-guided summarization variant achieved a BERTScore Recall of 0.5723 and a UMLS concept F1 of 0.1242, both lower than those of the best direct-generation configuration in our study (\autoref{tab:8b_results}). Similarly, no systematic gains were observed in ROUGE-1 or BLEURT. We attribute this behavior to the compressive nature of summarization: during abstraction, medically salient local findings and uncertainty expressions present in the candidates may be omitted, which is detrimental in medical captioning. Moreover, summarization is a rephrasing operation rather than a selection mechanism; instead of correcting factual errors introduced during generation, it may introduce additional errors by altering phrasing and inadvertently changing clinical meaning. Detailed results are reported in \autoref{tab:appen}.

\section*{Acknowledgment}

This work was supported by the National Research Foundation of Korea(NRF) grant funded by the Korea government(MSIT)(RS-2024-00360176).

\bibliographystyle{ieeetr}
\bibliography{reference}

\end{document}